%% file: main.tex
\documentclass[letterpaper]{article} 
\usepackage{aaai25}  
\usepackage{times}  
\usepackage{helvet}  
\usepackage{courier}  
\usepackage[hyphens]{url}  
\usepackage{graphicx} 
\usepackage{natbib}  
\usepackage{caption} 
\usepackage{algorithm}
\usepackage{algorithmic}

\usepackage{newfloat}
\usepackage{listings}
\DeclareCaptionStyle{ruled}{labelfont=normalfont,labelsep=colon,strut=off} 
\floatstyle{ruled}
\newfloat{listing}{tb}{lst}{}
\floatname{listing}{Listing}
\usepackage{tabularx}
\usepackage{subcaption}
\usepackage{multirow}
\usepackage{booktabs, makecell}
\usepackage{amsmath}

\usepackage[multiple]{footmisc}

\usepackage{xcolor}

\definecolor{green}{HTML}{008015}

\title{Privacy in Image Datasets: A Case Study on Pregnancy Ultrasounds}
\author {
    Rawisara Lohanimit\textsuperscript{\rm 1},
    Yankun Wu\textsuperscript{\rm 2},
    Amelia Katirai\textsuperscript{\rm 3},
    Yuta Nakashima\textsuperscript{\rm 2},
    Noa Garcia\textsuperscript{\rm 2}
}
\affiliations {
    \textsuperscript{\rm 1}Massachusetts Institute of Technology\\
    \textsuperscript{\rm 2}The University of Osaka\\
    \textsuperscript{\rm 3}Tsukuba University\\
    rloha@mit.edu,
    \{yankun@is., n-yuta@, noagarcia@\}ids.osaka-u.ac.jp,
    katirai.amelia.fu@u.tsukuba.ac.jp
}

\usepackage{bibentry}

\begin{document}

\maketitle

\begin{abstract}
The rise of generative models has led to increased use of large-scale datasets collected from the internet, often with minimal or no data curation. This raises concerns about the inclusion of sensitive or private information. In this work, we explore the presence of pregnancy ultrasound images, which contain sensitive personal information and are often shared online. Through a systematic examination of LAION-400M dataset using CLIP embedding similarity, we retrieve images containing pregnancy ultrasound and detect thousands of entities of private information such as names and locations. Our findings reveal that multiple images have high-risk information that could enable re-identification or impersonation. We conclude with recommended practices for dataset curation, data privacy, and ethical use of public image datasets.
\end{abstract}

%
\maketitle

\input{sec/1_introduction}
\input{sec/2_related_work}
\input{sec/3_methodology}
\input{sec/4_results}
\input{sec/5_conclusion}
\input{sec/6_ethic_statement}
\input{sec/7_adverse_impact}

\input{sec/8_acknowledgement}
\bigskip

\bibliography{aaai25}

\end{document}

%% file: sec/1_introduction.tex
\textbf{\section{Introduction}}
\noindent
The development of image generation models \cite{rombach2022high,saharia2022photorealistic,ramesh2022hierarchical} has significantly increased the demand for publicly available visual data. Image datasets \cite{deng2009imagenet,lin2014microsoft,schuhmann2021laion} have become essential for training robust computer vision models across multiple tasks, including image generation, image recognition, object detection, and others. 
However, the process of compiling large-scale image datasets is heavily based on web scraping techniques, collecting images from a wide variety of online sources, such as social media platforms, websites, and public repositories. Although scraping images from the internet allows one to collect massive amounts of data, it typically results in uncurated collections that might not meet quality or ethical standards, causing several critical issues. For example, image datasets can contain harmful societal biases \cite{birhane2021multimodal,meister2023gender,garcia2023uncurated}, such as reinforcing stereotypes based on race, gender, or age. Furthermore, uncurated datasets can also include sensitive or problematic content like child sexual abuse material \cite{caetano2025neglected, thiel2023identifying}, hate speeches \cite{birhane2024into}, or pornography \cite{birhane2021large}, which can lead to misuse of creating harmful outputs, such as synthetic child exploitation content \cite{thiel2023identifying,caetano2025neglected}. Another major concern is the potential exposure of individuals' personal information \cite{carlini2023extracting}, as some images may contain private or sensitive information that may be used without proper consent. 

When computer vision models are trained on such data, there is a significant risk that the undesirable traits, such as biases, harmful content, or privacy violations, are perpetuated or even amplified, potentially placing affected individuals and groups at risk \cite{katirai2024situating}. In the context of image generation, recent studies have highlighted that diffusion-based models can memorize training examples,  producing exact or near-identical copies of images during the generation process \cite{carlini2023extracting}. This becomes particularly critical if the training dataset contains personal or sensitive data, as these generative models could potentially reproduce private information, allowing anyone to access it. To address these issues, the field of machine unlearning \cite{gandikota2023erasing,Zhang_2024_forgetmenot,lu2024mace} proposes algorithmic solutions to steer models away from generating unwanted concepts. However, there is evidence \cite{suriyakumar2024unstable} showing that models are not able to fully forget a concept learned during training. To directly address the root causes of these issues, other efforts are focused on auditing datasets \cite{birhane2021large,birhane2021multimodal,garcia2023uncurated,thiel2023identifying,birhane2024into}. For example, \citet{birhane2024into} audit LAION-400M \citep{schuhmann2021laion} and LAION-2B, two subsets of the full LAION-5B \cite{schuhmann2022laion}, revealing an increasing prevalence of hate and toxic content as dataset sizes grow. 

Inspired by previous works on dataset auditing, this paper aims to investigate privacy concerns in large-scale image datasets. As privacy in computer vision refers to a broad term that can encompass various types of sensitive information in visual data, ranging from recognizable facial features to medical information, we narrow our focus by investigating pregnancy ultrasound images. Pregnancy ultrasound images, or sonograms, are medical images produced using ultrasound technology to visualize a developing fetus in the womb. They are taken by using high-frequency sound waves to create real-time images of the fetus, placenta, and surrounding tissues with the aim of monitoring fetal growth, detecting any potential abnormalities or health concerns, and determining important factors such as gestational age. Pregnancy ultrasounds can also provide visual information about the pregnant person’s reproductive organs and the overall health of the pregnancy. While previous works have uncovered the existence of misogyny, pornography, malignant stereotypes \cite{birhane2021large,birhane2021multimodal,birhane2024into}, medical images \cite{adams2023does} and child sexual abuse material (CSAM) \cite{thiel2023identifying} in large uncurated image datasets, the choice of pregnancy ultrasounds as our case study is motivated by the unique characteristics of these images:

\begin{itemize}
    \item Pregnancy ultrasounds often contain a significant amount of personal information, such as the name of the pregnant person, the hospital where the ultrasound was performed, the date and time of the procedure, the gestational age of the fetus, or the name of the sonographer (the person performing the ultrasound). This personal information can be highly sensitive and potentially identifiable \cite{leaver2018visualising}. 
    
    \item Pregnancy ultrasounds are frequently shared online or on social media platforms as part of pregnancy announcements to family and friends, increasing the likelihood to be collected in image datasets used to train computer vision models and posing a substantial risk to privacy. 
        
    \item Pregnancy ultrasounds can reveal potential malformations or health issues in the fetus and the pregnant person, such as developmental anomalies or genetic conditions, which are highly sensitive medical information. If such images are included in a dataset and leaked, they could expose private details about the health and future well-being of both the pregnant person and the fetus. 

    \item The intimate nature of pregnancy ultrasounds makes them highly personal, and their inclusion in training datasets without consent can lead to unintended harm or emotional distress for the individuals involved, especially if these images are used in ways that the individual did not anticipate or approve. 

    \item Ultrasound images are often taken in medical settings where the pregnant person is vulnerable. The improper use or public exposure of such images could cause distress, erode trust in healthcare systems, and further contribute to the broader societal issues on data privacy. 
\end{itemize}

Overall, pregnancy ultrasound images have the potential to contain highly sensitive private information, while at the same time, they are frequently shared and celebrated online, increasing the likelihood to be included in image datasets. Thus, our work seeks to answer the following two key research questions (RQs): 
\begin{enumerate}
    \item Are pregnancy ultrasound images present in image datasets used to train computer vision models?
    \item If so, do pregnancy ultrasound images contain private information that can be used to identify individuals? 
\end{enumerate}

To address RQ(1), we develop a method for detecting pregnancy ultrasound images based on large-scale image retrieval and classification techniques and apply it to the LAION-400M dataset \citep{schuhmann2021laion}. Using this approach, we successfully identify $833$ pregnancy ultrasound images, which we further analyze to respond to RQ(2). We run a personal information detection algorithm based on OCR text recognition and Named Entity Recognition on these identified images and uncover four types of personal information: name, location, date time, and phone numbers. 
While the number of detected images may seem small relative to the 400 million images originally in the dataset, it is important to emphasize that these are real individuals, with real pregnancies, and real lives at stake. The inclusion of such images in the dataset without consent is ethically problematic and must be addressed with great care. 
Thus, we conclude the paper with several recommendations for large-scale image collections, including the implementation of more robust data privacy and consent protocols, especially for images containing sensitive personal information.

%% file: sec/2_related_work.tex
\section{Background}

\subsection{Private Information Definition}
There are varied approaches and concepts related to private information and the question of how to define privacy and what should be considered to be private is often contentious, even among scholars in the area. Legal definitions of privacy vary depending on regional or cultural context, as evidenced, for example, by the overlapping but distinct definitions of personally identifiable information in the United States,\footnote{\url{https://www.dol.gov/general/ppii}} and personal or sensitive data in the European Union.\footnote{\url{https://commission.europa.eu/law/law-topic/data-protection/rules-business-and-organisations/legal-grounds-processing-data/sensitive-data/what-personal-data-considered-sensitive_en}}
As online ultrasound sharing has been reported in the academic literature across multiple regional and cultural contexts \cite{zhu2019pregnancy,Roberts2015WhyDW}, and working as an interdisciplinary and international team, we avoid a regionally-bound legal definition of privacy and instead use the broader term ``private information'' and an approach informed by recent academic debates on privacy.

Many key privacy scholars argue for an anti-reductionist, pluralistic understanding of privacy which recognizes that privacy is interlinked with too many aspects, including specific circumstances or contexts, to be able to be reduced to a single definition \cite{trepte2023definitions}. Moreover, as \citet{solove2023data} has argued in relation to sensitive data, ``the borderlines of many categories are so blurry that they are useless,'' and that, ultimately, ``data is what data does''.

Acknowledging these debates and varied concepts, we utilize Nissenbaum's theory of contextual integrity here, \cite{nissenbaum2004privacy,malkin2022contextual}, which highlights the importance of norms and expectations for how information will be used within particular contexts, and the resulting privacy violations that occur when these expectations are breached and data is utilized outside of these expected ways. Thus, \emph{data which may not be too sensitive to share in one context, may become so when that data is removed from its original context and utilized for another purpose, as in the case of data scraped for training.} This is especially relevant in the area of pregnancy ultrasounds, where an ultrasound image may be shared to mark a particular social and medical milestone on an online platform where the poster expects that friends and family or other interested people may be able to view it, but a breach of contextual integrity occurs when this data is extracted from this intended context and used as a part of training datasets.

It is also noteworthy that unintended access or use to health-related data such as ultrasounds can create risks related to insurance or identity theft, and are frequently highly sensitive.\footnote{
\url{https://www.propublica.org/article/millions-of-americans-medical-images-and-data-are-available-on-the-internet}}\footnote{\url{https://techcrunch.com/2020/01/10/medical-images-exposed-pacs/}} Pregnancy ultrasounds in particular can reveal sensitive information about parental or fetal health and healthcare access, and so usage of the images beyond their original intended scope can be highly problematic.\footnote{\url{https://www.curtin.edu.au/news/think-before-you-post-the-impact-of-sharing-photos-of-your-child-online/}}

\subsection{Privacy Information in Images and in Datasets}
With the growing concerns about the presence of private information in images, researchers have developed privacy detection methods to determine whether an image contains private information \cite{10.1145/3386082, tran2016privacy, tonge2016image}. Typical approaches extract features using a trained model and use them to train a classifier that predicts the presence of private information~\cite{xu2024examining, zhao2022privacyalert}.  A more complex task involves identifying specific areas within an image that contain private information \cite{gurari2019vizwiz}.

The focus of image privacy often revolves around bioidentifiers, such as facial attributes \cite{zhang2014anonymous, kumawat2022privacy, carlini2023extracting}. For example, \citet{zhang2014anonymous} proposed an anonymous camera that uses optical masking to obscure captured faces, thereby improving privacy during image collection.
A recent study~\cite{samson2024privacy} finetuned a visual language model to improve the private information detection ability using a dataset constructed from LAION-5B. While this improved detection capabilities, it also raised concerns about the potential presence of private information in large-scale image datasets and the risks they pose.

Despite these advancements, the issue of private information presence within image datasets remains insufficiently explored compared to other types of datasets. For example, \citet{LI2024101916} evaluated the privacy and re-identification risks of open government data in China. In addition, millions of instances of private data, such as email addresses, phone numbers, IP addresses, credit card numbers, bank account numbers, and names, have been found across large text corpora \cite{subramani2023detecting, elazar2023s, jahan2023analysis} such as the Colossal Clean Crawled Corpus \cite{raffel2023exploringlimitstransferlearning} and the Pile \cite{gao2020pile800gbdatasetdiverse}. In contrast, despite progress in studying other issues in image datasets, such as toxic content \cite{birhane2024into}, copyright concerns \cite{ma2024dataset, moayeri2024rethinking}, metadata anonymization \cite{rempe2024identification, jahan2023analysis}, and demographic bias \cite{garcia2023uncurated, meister2023gender}, privacy risks have not been widely evaluated. This gap creates potential risks of private information leakage \cite{hu2024vlsbench, zhang2022differential}. Unlike medical datasets, which often implement privacy-preserving measures to remove sensitive data before publication \cite{zhang2022differential}, such practices are rarely applied to image datasets from other domains \cite{deng2009imagenet, lin2014microsoft}, leaving them vulnerable to the exposure of private information. Co-current work by \citep{hong2025common} investigates general privacy concerns in a public image dataset, DataComp CommonPool \cite{gadre2023datacomp}, with a primary focus on the legal implications and the effect of data filtering. In contrast, our work studies pregnancy ultrasound images, analyzing subtypes and co-occurring private information that heighten re-identification risks. 
\begin{center}
    \begin{figure*}[t]
    \centering
  \includegraphics[width=0.66\linewidth]{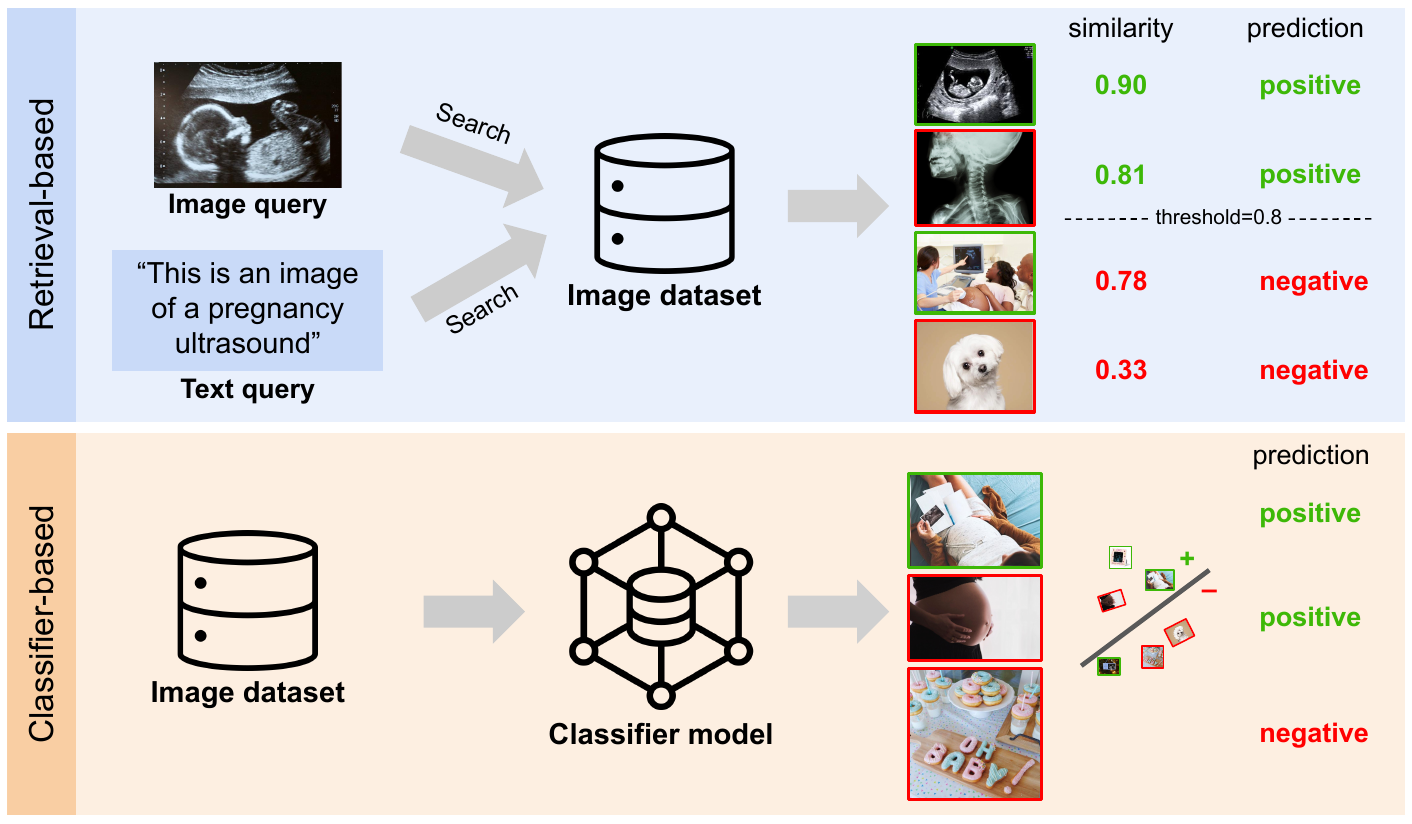}
    \vspace{-5pt}
  \caption {\label{fig:search} Pregnancy ultrasound image detection. For retrieval-based detection, we use image and text as queries to find images that have high similarity with the query. For classifier-based detection, we use the output of the classifier as the prediction.}
\end{figure*}
    \vspace{-8pt}

\end{center}

\vspace{-10pt}
\subsection{The Pregnancy Ultrasound Image}
Next, we briefly review prior research on the significance of ultrasound images, finding that their significance is not only medical, but also social, and is shaped by culture \cite{Roberts2015WhyDW}. 

First, pregnancy ultrasounds carry medical significance as they are perceived to be a part of appropriate antenatal care, and as necessary for a successful pregnancy \cite{oyen2016viewing}. This is in part due to their use in diagnosing congenital conditions. However, there are also concerns over the role of ultrasounds in the medicalization of pregnancy, through which the pregnant person increasingly comes to be viewed as a patient, and pregnancy as a dangerous state, which must be controlled and monitored through technology. In addition, ultrasounds carry social significance as one method through which both parental and fetal identities are developed. Though there is a belief that ultrasounds contribute to parental bonding with a fetus \cite{skelton2024influence}, this is contested in recent research. It is also noteworthy that the cultural weight of ultrasound images has been critically examined by feminist scholars, who argue that ultrasound images and their role in identity construction constrain the right of the pregnant person to choose \cite{lie2019he}.

The social significance is related to cultural pressures, including pressures to save the images ``for posterity,'' and to share them with others, which increasingly occurs through online platforms \cite{HARPEL2018sharing}. As such, the ultrasound has been described as part of an ``online birth'' taking place prior to the ``physical birth'' \cite{johnson2014maternal} as it becomes part of a digital footprint which extends across the life-course. Such images are primarily shared with family or friends, as one study of $117$ pregnant women conducted through Facebook found that $77.6$ percent shared only with friends, and $78.4$ with family \cite{HARPEL2018sharing}. Similarly, a study of Chinese expectant mothers found that pregnancy was perceived to be a time of vulnerability \cite{zhu2019pregnancy} and thus news regarding it was shared primarily within closed groups of family or friends. However, they may also be shared more broadly, as \citet{leaver2018visualising} report the presence of ultrasounds publicly shared through Instagram . Further, their study identified privacy concerns including the presence of what the authors termed personally identifiable metadata. These issues are then further amplified when such images are scraped and used for purposes unanticipated by the original sharer. 

%% file: sec/3_methodology.tex
\section{Methodology}
\label{sec:methodology}

To address our research questions, RQ(1) and RQ(2), we develop a methodology that aims to: 1) detect pregnancy ultrasound images within large collections of images, and 2) identify private information in pregnant ultrasound images. 

\subsection{Pregnancy Ultrasound Image Detection}
\label{sec:ultrasound_detection}
Given a dataset of images $\mathcal{D} = \{I_1, I_2, \cdots, I_N\}$ where $I_j$ represents the $j$-th image in a dataset of size $N$, the goal is to identify the subset $\mathcal{U} \subset \mathcal{D}$ such that images in $\mathcal{U}$ contain pregnancy ultrasound images. To achieve this, we follow two approaches: a retrieval-based approach and a classifier-based approach. An overview is shown in Figure \ref{fig:search}.

In both approaches, images in $\mathcal{D}$ are mapped to a semantic space with a pre-trained CLIP image encoder. The retrieval-based approach uses a query, which can be either a text description or an example ultrasound image, to compute the similarity between images in $\mathcal{D}$ and the queries. An image in $\mathcal{D}$ is identified as an ultrasound image when the similarity is sufficiently large. In contrast, the classifier-based approach uses a classifier trained on a training dataset containing both pregnancy images as positive samples and other images as negative samples.

\subsubsection*{Retrieval-Based Approach} 
This approach uses a set $\mathcal{Q}_\text{t} = \{Q_\text{t}\}$ of text description queries $Q_\text{t}$ or a set $\mathcal{Q}_\text{i} = \{Q_\text{i}\}$ of image queries $Q_\text{i}$ that exemplify ultrasound images. We use pre-trained CLIP text and image encoders\footnote{\url{https://github.com/rom1504/clip-retrieval}} to map the queries and images in $\mathcal{D}$ into the semantic space. Let $\text{CLIP}_{\text{image}}$ and $\text{CLIP}_\text{text}$ denote the CLIP image and text encoders, respectively. We compute image features $x_j = \text{CLIP}_{\text{image}}(I_j)$ for $I_j \in \mathcal{D}$, as well as text features $q_\text{t} = \text{CLIP}_\text{text}(Q_\text{t})$ for $Q_\text{t} \in \mathcal{Q}_\text{t}$ and image features $q_\text{i} = \text{CLIP}_\text{image}(Q_\text{i})$ for $Q_\text{i} \in \mathcal{Q}_\text{i}$. Some example text queries $\mathcal{Q}_\text{t}$ are: \textit{Ultrasound imaging for prenatal screening}, \textit{Pregnancy announcement photoshoot}, and \textit{Prenatal ultrasound for comprehensive fetal anomaly screening}, which cover different types of possible pregnancy ultrasound images.

We use $|\mathcal{Q}_\text{i}| = 22$ images and $|\mathcal{Q}_\text{t}| = 100$ sentences as queries. We compute cosine similarity between the representations of queries and images in $\mathcal{D}$.
For each image in $\mathcal{D}$, we identify the highest similarity among all similarities computed with the queries. If the highest similarity exceeds the threshold $\tau$, the image is retrieved as a pregnancy ultrasound image.

\subsubsection*{Classifier-Based Approach} This approach classifies whether $I_j$ is a pregnancy ultrasound image using CLIP image features $x_j$. We compare three different methods: random forest (RF), support vector machine (SVM), and neural network (NN). We use scikit-learn \citep{scikit-learn} implementation of RF and SVM classifiers, while the NN model is implemented with PyTorch \citep{pytorch}. The parameters of the RF and SVM models are optimized using grid search. The NN model consists of four layers with dimensionalities $1,024$, $256$, $32$, and $1$, respectively, and ReLU activations \cite{agarap2018deep}. The binary cross entropy loss function is used with Adam optimizer \cite{kingma2014adam}, a learning rate of $0.0001$, and a batch size of $1,024$. 

\begin{figure*}[t]
    \centering
    \begin{subfigure}{0.43\textwidth}
        \centering
        \includegraphics[width=0.98\linewidth]{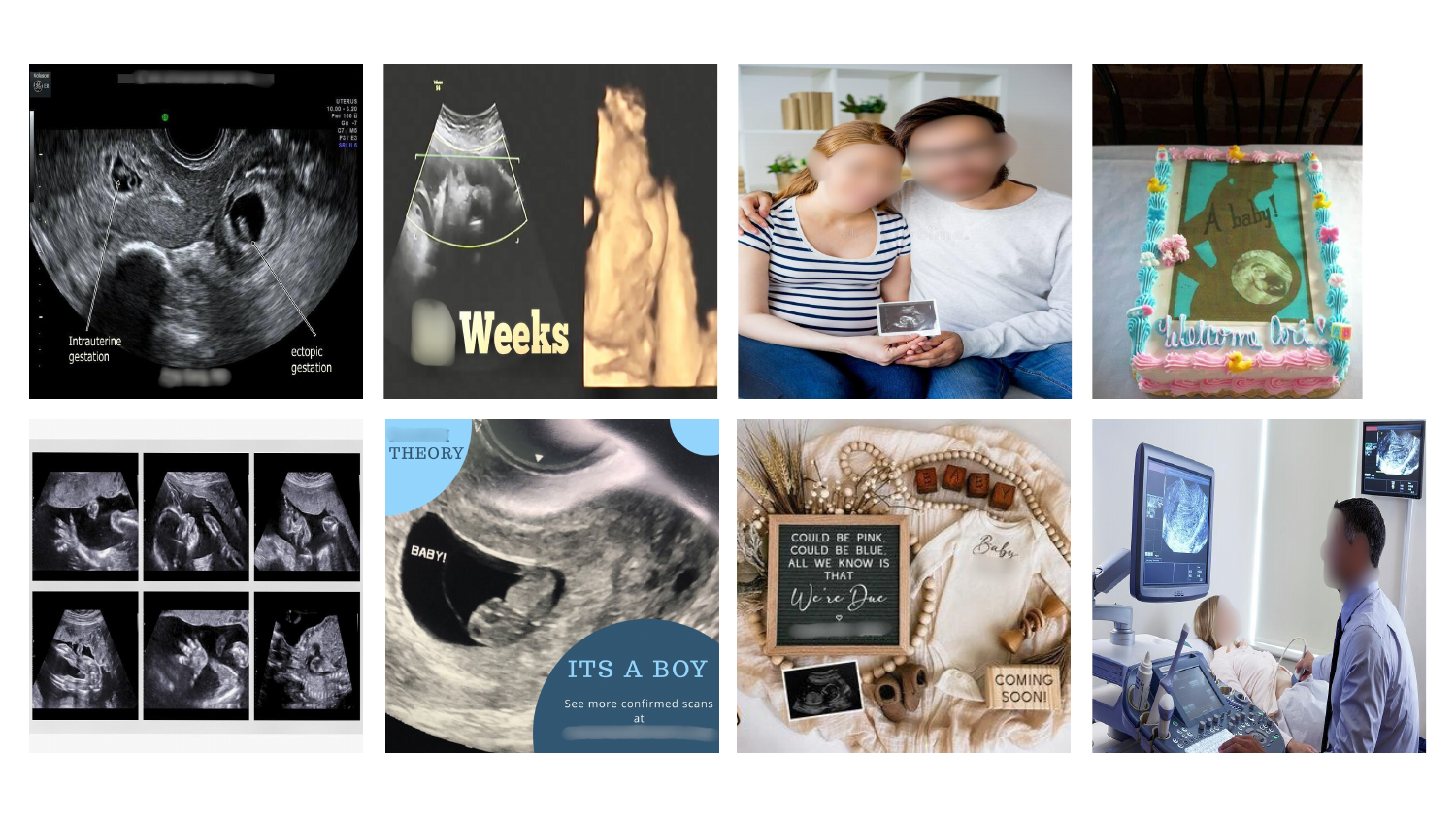}
        \caption{\label{fig:PUI_ultrasound}pregnancy ultrasound images}
    \end{subfigure}
    \hfill 
    \begin{subfigure}{0.43\textwidth}
        \centering
        \includegraphics[width=0.98\linewidth]{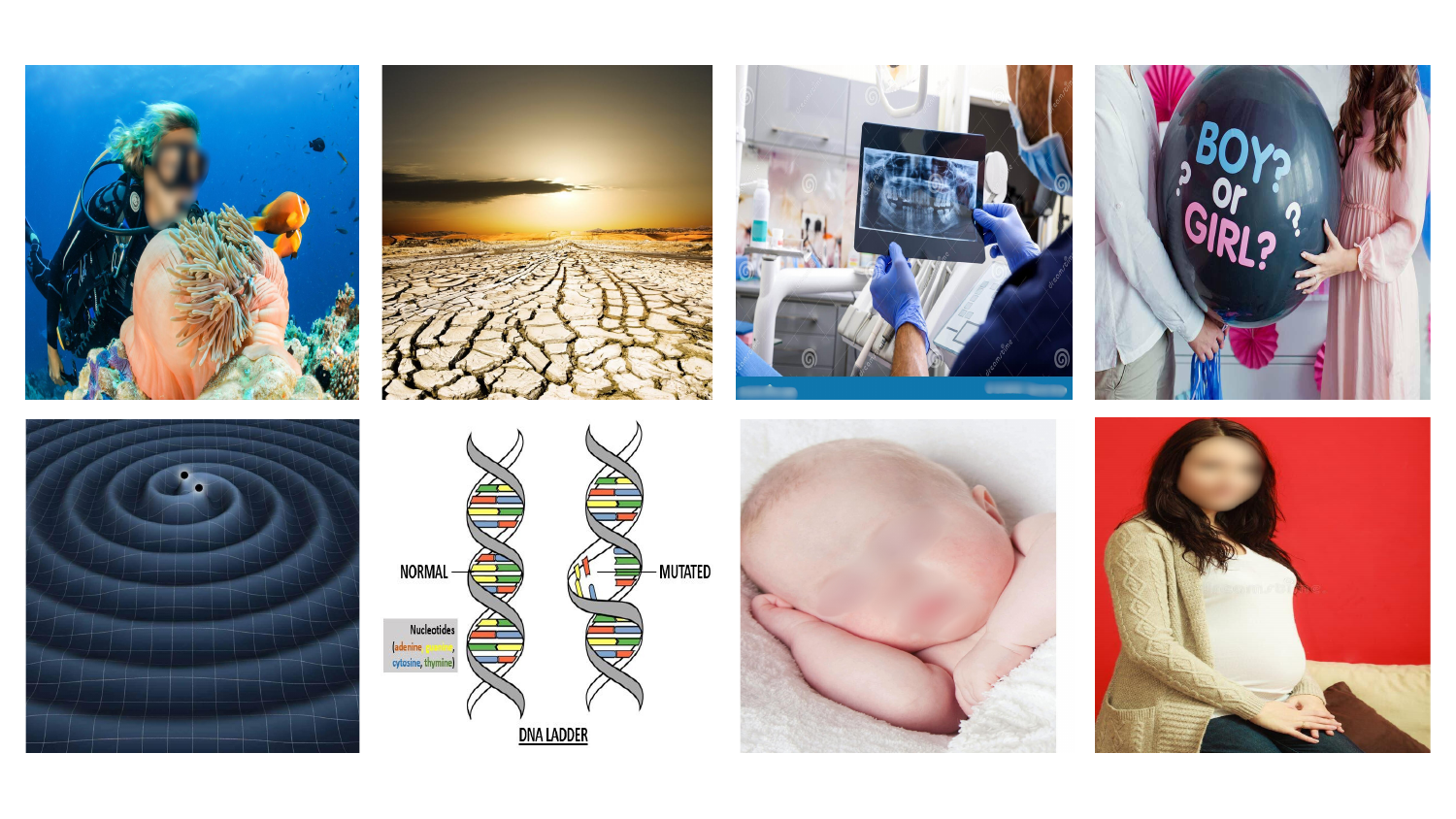}
        \caption{\label{fig:PUI_nonultrasound} non-pregnancy ultrasound images}
    \end{subfigure}
        \vspace{-8pt}

    \caption{\label{fig:PUI_samples}Examples of (a) positive images and (b) negative images in the PIU dataset. Faces and private information redacted for privacy.}
    \vspace{-5pt}

\end{figure*}

\subsubsection*{The PIU Dataset}
To train the parameters of the classifiers, we collect the PIU (Pregnancy Image Ultrasound) dataset. PIU is a custom-made dataset with both positive and negative examples of pregnancy ultrasound images collected by using the easy-image-scraping library \citep{naumannScrapeCutPasteLearn2022}. Due to the emotional significance of pregnancy ultrasound images for expectant parents, these images appear in various forms when shared online. Common presentations include plain digital ultrasound images, photos of ultrasounds with decorative backgrounds, ultrasounds incorporated into event invitations such as baby showers, and family photos featuring members holding ultrasound images. To account for all this variability and get a sufficient representation of ultrasound images for training our classifiers, we query images using keywords such as \textit{pregnancy ultrasounds}, \textit{baby shower with ultrasounds}, and more. For negative examples, on top of collecting any non-related instances such as \textit{animals}, \textit{locations}, or \textit{household objects}, we especially look for images under the concepts of \textit{non-pregnancy ultrasound}, which tend to be the hardest cases to classify. 

We split a preliminary collection of images into train, validation, and test splits. Using an active learning approach, we iteratively refine the training set while keeping the validation and test sets fixed. In each iteration, we train a model, evaluate it on the validation dataset, and identify misclassified or challenging images. Then, we collect similar images to add to the training set. To prevent data leakage, we remove images with CLIP similarity above 0.95 to those in the validation or test sets both before initiating active learning and during each iteration, followed by manual inspection. The model is retrained with the updated train split. We repeat this process until the validation performance stabilizes, allowing the model to adapt to diverse data in large image datasets without overfitting. The final dataset contains 9,900 images, evenly balanced between positive and negative, comprising 3,960 training, 990 validation, and 990 test images. Figure \ref{fig:PUI_samples} shows example images.

\subsection{Private Information Identification}
\label{sec:information_detection}
\begin{center}
    \begin{figure*}[t]
    \centering
  \includegraphics[width=0.73\linewidth]{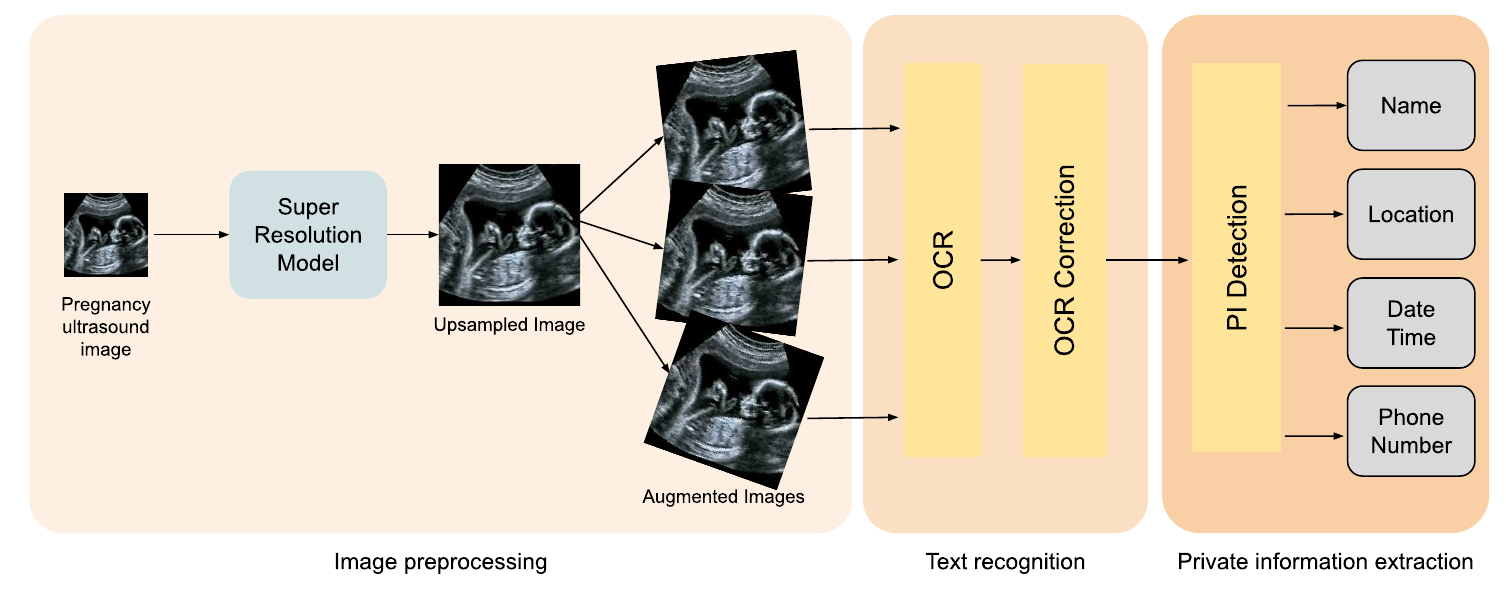}
  \caption {\label{fig:private_info} Private information identification. Detected pregnancy ultrasound images are 1) preprocessed with super-resolution and rotation for horizontal text alignment, 2) processed for text recognition and correction, and 3) passed to a private information detection system to extract Name, Location, Date Time, and Phone Number entities.}
\end{figure*}
\end{center}

After detecting pregnancy ultrasound images in the original dataset $\mathcal{D}$, the next step is to search for private information within them. We develop an approach (see Figure \ref{fig:private_info}) that detects and reads text in images in three steps: image preprocessing, text recognition, and private information extraction.
\subsubsection*{Image Preprocessing} 
Since pregnancy ultrasound images can appear in various formats and visual presentations, and due to the uncurated nature of large image datasets, text in the images may not always be clearly readable by algorithms. Challenges such as handwritten text, image tilt, and low resolution can hinder text detection. To address these issues, we apply image preprocessing. First, we upscale low-resolution images (i.e. one of the dimensions is less than 200 pixels) by a factor of four using the Real-ESRGAN super-resolution model \citep{wang2021realesrgan}. Additionally, we augment each image by rotating it by 5 to 90 degrees, both clockwise and counterclockwise. These image preprocessing steps improve text detectability in subsequent stages.
\vspace{-18pt}
\subsubsection*{Text Recognition} Next, we employ an open-source optical character recognition (OCR) model called Tesseract \cite{kay2007tesseract} to read text in images. Tesseract OCR is an LSTM-based model \citep{hochereiter1997lstm} that recognizes line and character patterns in images. However, despite the image preprocessing step, the initial OCR outputs are not always accurate. To refine the quality of the recognized text, we employ the LLaVa-Next model \citep{liu2023llava, liu2023improvedllava}, a general-purpose visual and language understanding model. LLaVa-Next can correct OCR errors by conditioning on the image with the prompt: \textit{Correct the following OCR extracted text: [recognized text].}

\subsubsection*{Private Information Extraction} From the recognized text in the previous step, 
we utilize Presidio \citep{mendels2018presidio} to identify private information. Presidio is a software development kit capable of identifying and anonymizing private entities in text, such as credit card numbers, names, locations, and more. It leverages natural language processing techniques, including regular expressions for pattern matching, Named Entity Recognition to detect entities, and rule-based logic and checksum with relevant context. The software has both predefined and customizable recognizers to detect sensitive entities. Following \citet{subramani2023detecting}, we consider various types of private information and narrow them down to four categories — Name, Location, Date Time, and Phone Number — based on their frequent occurrence in the images.\footnote{Detailed definitions of types and recognizers can be found in the appendix.}

%% file: sec/4_results.tex
\section{Private Information in LAION-400M}
Following the methodology described above, we examine the LAION-400M dataset \cite{schuhmann2021laion} in search for private information, specifically private information in pregnancy ultrasound images. LAION-400M is a dataset with $400$ million image-text pairs in English extracted from the Common Crawl web data dump between 2014 and 2021 and filtered with CLIP \cite{radford2021learning} to remove non-matching semantic text-image data. We select this dataset due to its public availability and its widespread adoption in training vision-language models, such as the popular Stable Diffusion \citep{rombach2022high}. We use the precomputed CLIP image embeddings as image representations. Our findings are presented below.

\begin{center}
    \begin{figure*}[h!]
    \centering
  \includegraphics[width=0.58\linewidth]{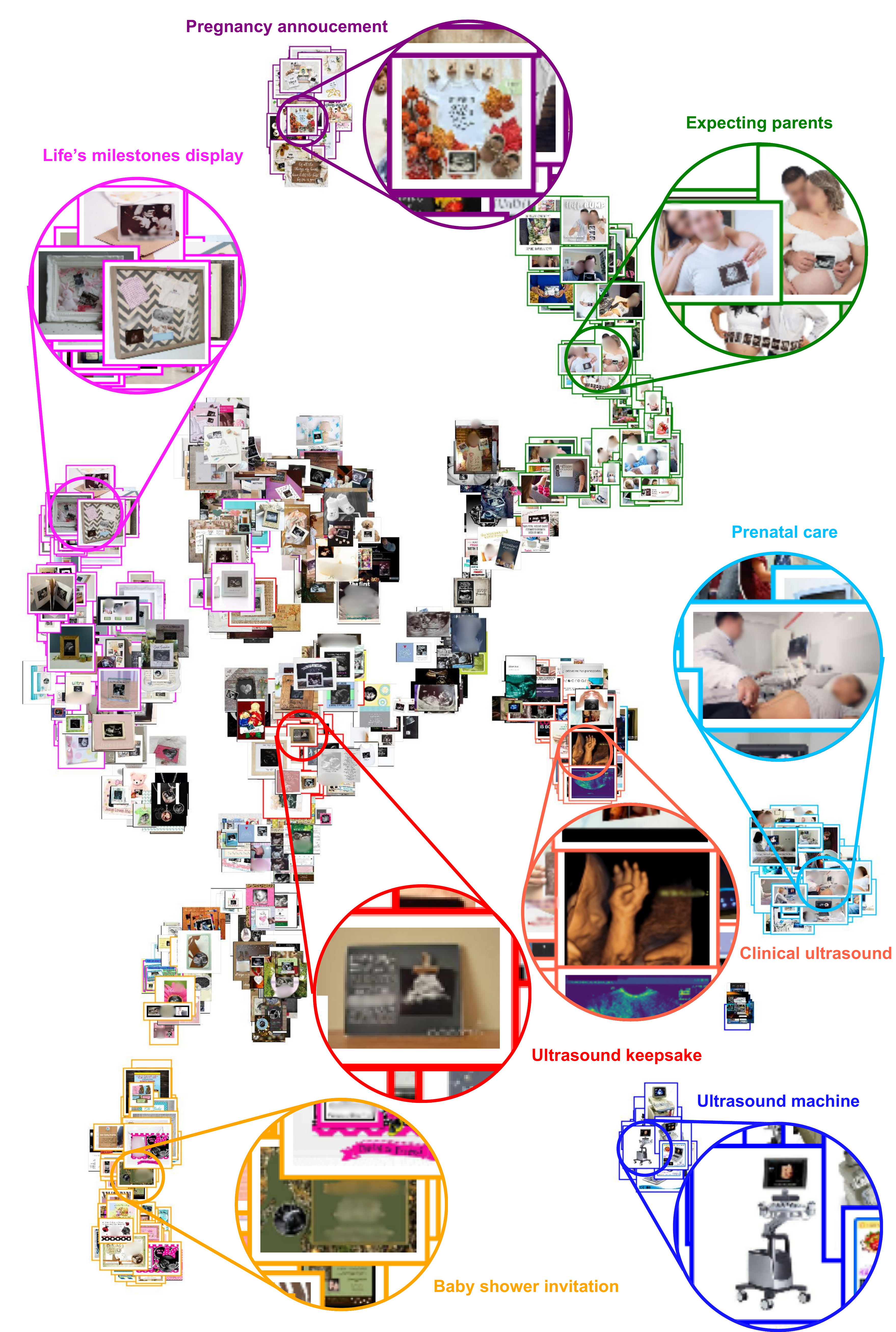}
  \caption {\label{fig:cloud_plot} t-SNE visualization of the pregnancy ultrasound images found in LAION-400M. Colors represent each of the cluster themes (names shown next to each cluster) found with HDBSCAN. Faces and private information redacted for privacy.}
\end{figure*}
\end{center}
\subsection{Pregnancy Ultrasound Images in LAION-400M}

\paragraph{Search Method Evaluation}

First, we evaluate the detection performance of the two approaches, i.e. the retrieval-based detection and the classifier-based detection, on the PIU dataset. For the retrieval-based approach, we select an optimal threshold on the PIU validation set, and then apply the chosen threshold to compute results on the test set. When images are used as queries (i.e., image-to-image retrieval), the threshold is set at $\tau=0.7$, while for text queries (i.e., text-to-image retrieval), $\tau=0.3$.

The results are presented in Table \ref{tab:model_performance}. With retrieval-based detection, despite achieving high similarity scores, the accuracy in detecting pregnancy ultrasound images is relatively low, particularly when using images as references. This could be attributed to the high-level semantics in the CLIP embedding space. For example, images of babies or pregnant individuals have a high similarity score with pregnancy ultrasound images in the CLIP embedding space, leading to the lower accuracy of the retrieval of \textit{pregnancy ultrasound images} specifically and reducing retrieval accuracy. This indicates that more specialized models are needed to accurately identify pregnancy ultrasound images. In contrast, the classifier-based detection, which trains a dedicated classifier on top of the CLIP embeddings using PIU training images, obtains much better performance. From the three proposed classifiers, the SVM  stands out for its high accuracy and low false positive rate. This is particularly important given the large volume of data; even a small false positive rate can lead to a significant number of irrelevant results that require manual inspection when applied to a large dataset. The SVM’s false negative rate is also comparable to the best performing classifier, which is crucial for ensuring that target images are not missed. Therefore, for the rest of the paper, we present results using the SVM classifier.

\begin{table}[t]
\footnotesize
\renewcommand{\arraystretch}{1}
\setlength{\tabcolsep}{10pt}
    \centering
    
    \begin{tabularx}{0.45\textwidth}{@{}Xl@{\hspace{0.8em}}r@{\hspace{0.8em}}r@{\hspace{0.8em}}r@{}}
        \toprule
        \textbf{Detection approach} & \textbf{Method} & \textbf{Acc} & \textbf{FP rate} & \textbf{FN rate} \\ 
        \midrule
        \multirow{2}{*}{Retrieval-based} & Image & $77.17$ & $10.55$ & $34.75$ \\
        & Text & $83.03$ & $8.52$ & $25.05$ \\
        \hline
        \multirow{3}{*}{Classifier-based} & RF & $94.64$ & $5.05$ & $5.65$ \\
        & SVM & $\textbf{97.27}$ & $\textbf{1.41}$ & $4.04$ \\
        & NN & $96.67$ & $3.23$ & $\textbf{3.43}$ \\ 
        \bottomrule
    \end{tabularx}
    \caption{Pregnancy ultrasound image detection performance on the PIU test set.}
    \label{tab:model_performance}
    \vspace{-8pt}
\end{table}

\paragraph{Detected Pregnancy Ultrasound Images in LAION-400M}
Using the trained SVM model on the whole LAION-400M dataset, we identify $1,364$ pregnancy ultrasound images. 
To remove duplicates, we use a copy detection method SSCD \cite{pizzi2022self} with a similarity threshold of $0.92$, and keep one image per duplicate set. This process reduces the image count to $1,045$. We subsequently apply manual inspection, further refining it to $833$ unique images, highlighting the significant presence of duplicate data within the dataset, as previously noted by \cite{touvron2023llama}. 

\paragraph{False Positive Rate} To validate the efficiency of our model on the target dataset, we inspect potential false negative images. As finding actual false negatives among 400 million unlabeled images is highly challenging, we conducted a manual inspection of images near the classifier’s decision boundary. Specifically, we applied the SVM classifier to a partition of the LAION-400M dataset (i.e. 1 million images) and manually reviewed images whose negative prediction scores fell within 2 standard deviations (S.D.) of the decision boundary. We selected this threshold because only 38 images were within 1 S.D., making it insufficient for a meaningful analysis. Within the 2 S.D. range, we identified 1,872 images, none of which were pregnancy ultrasound images.
 
\begin{table*}[t]
\footnotesize
\setlength{\tabcolsep}{10pt}
\renewcommand{\arraystretch}{1}
    \centering
    \begin{tabular}{lllr}
        \toprule
        \textbf{ID} & \textbf{Theme} &\textbf{Most frequent words} & \textbf{Num. images}\\ \midrule
        0&Ultrasound machine & ultrasound, doppler, color, scanner, machine &  33 \\
        1&Prenatal care& ultrasound, pregnant, woman, stock, doctor & 55\\
        2&Expecting parents& pregnant, pregnancy, stock, baby, ultrasound  & 120 \\
        3&Clinical ultrasound& ultrasound, fetal, fetus, pregnancy, abortion & 33 \\
        4&Pregnancy announcement& announcement, pregnancy, baby, social, media & 40 \\
        5&Baby shower invitation& baby, shower, invitations, invitation, photo & 67 \\
        6&Ultrasound keepsake& frame, photo, baby, picture, sonogram & 27 \\
        7&Life's milestones display& baby, frame, photo, box, shower& 82 \\
        -1& Not in any cluster & - & 376\\ 
        \bottomrule
        All & - & - &  833 \\
        \bottomrule
    \end{tabular}
    \caption{Cluster analysis of pregnancy ultrasound images detected in the LAION-400M dataset. Each image is assigned to a cluster based on HDBSCAN results. To characterize the themes of each cluster, we extract the top 5 most frequent words from the captions of the images within each cluster and use them to assign theme names. }
    \label{tab:cluster_count}
    \vspace{-5pt}
\end{table*}
\paragraph{Data Visualization and Clustering}
We further analyze the different types of detected pregnancy ultrasound images using data visualization tools. First, we transform the CLIP image representation of each image from the original 512-dimensional space into a 5-dimensional vector using UMAP~\citep{mcinnes2020umap}. We then apply HDBSCAN \citep{McInnes2017hdbscan} for clustering, with a minimum cluster size of 20 images. This approach reveals 8 distinct clusters. In contrast, when HDBSCAN is applied directly to the original CLIP image representation without UMAP dimensionality reduction, no clusters are detected, highlighting the necessity of reducing the dimensionality to reduce information from the 512-dimensional space, which may otherwise introduce distracting noise.

Next, we visualize the 833 pregnancy ultrasound image space in two dimensions by reducing the image features to 2D using the t-SNE algorithm~\citep{vanDerMaaten2008tsne}. Figure \ref{fig:cloud_plot} shows the spatial arrangement of the clusters, revealing varying degrees of semantic relatedness, with some outliers such as images of ultrasound machines and pregnancy announcement photos. Despite efforts to eliminate duplicates, some similar images remain due to resolution differences, which caused the image representations to be less semantically aligned.

\paragraph{Pregnancy Ultrasound Images Themes}

When analyzing each of the 8 detected clusters in detail, we observe
unique themes across them. To effectively label these themes, we compute word frequency counts from the captions associated with the images in each cluster, as the captions are semantically aligned with the images they describe. The top 5 most frequent words for each cluster are used to manually set cluster names that capture the predominant themes in the images. Additionally, clusters are color-coded in the t-SNE representation map shown in Figure \ref{fig:cloud_plot}, providing an intuitive visualization of the types of pregnancy ultrasound images found in the LAION-400M dataset. A detailed breakdown, including the themes, the top 5 most common words, and the number of images per cluster, is provided in Table \ref{tab:cluster_count}.

While our initial expectation of the dataset's ultrasound images leaned towards the stereotypical medical film format, black and white with technical details like date, time, and hospital information, our analysis reveals a broader variety. Although these ideal information-rich images are still present in the dataset, they are a minority. Upon visual and semantic examination, none of the clusters distinctly represents this stereotypical format. The most prevalent category is \textit{expecting parents} (ID 3), which accounts for 14.4\% of the images, depicting women or couples holding ultrasound images. This aligns with real-world scenarios where expectant parents share their pregnancy journey online, using these images for announcements and to foster social engagement \citep{HARPEL2018sharing,Roberts2015WhyDW}. Furthermore, \textit{life's milestones display} (ID 7) represents 9.8\% of the images, emphasizing their sentimental value as keepsakes. Another notable category is \textit{baby shower invitation} (ID 5) accounting for 8\% of the images. While the actual ultrasound images are smaller, these images contain significant private information relevant to event details such as names, addresses, phone numbers, and dates. This presents a potential risk regarding data privacy as the inclusion of such detailed private information in publicly shared or inadequately secured images can lead to unauthorized access and misuse of this data. 

The majority of images did not belong to any cluster, potentially due to their unique characteristics or insufficient numbers to form a separate cluster. These images include various items such as pendants, decorations, and book covers. Although some of these images bear a superficial resemblance to existing clusters, their distinct visual features prevent them from being grouped with other more cohesive clusters. Moreover, another observation is the low density of these images in Figure \ref{fig:cloud_plot}, suggesting sparse representation in the feature space.

\subsection{Private Information in Pregnancy Ultrasound}
\begin{table}[t]
\small
\setlength{\tabcolsep}{10pt}
\renewcommand{\arraystretch}{1}

        \centering
        \vspace{-5pt}
        \begin{tabular}{lrr}
        \toprule
        & \multicolumn{2}{c}{\textbf{Num. instances}} \\ \cline{2-3}
        \textbf{Type} & \textbf{All images} & \textbf{Unique images}  \\  \hline
        Name &  $387$ & $228$\\
        Location & $238$ &  $120$ \\
        Date Time  & $513$  & $299$\\
        Phone Number & $55$ & $30$ \\
         \hline
        Total & $1,193$& $677$\\
         \hline
    \end{tabular}
     \caption{Summary of private information instances extracted from pregnancy ultrasound images in LAION-400M.}
     \label{tab:PI_count}
\end{table}

\begin{table}[t]
\small
\setlength{\tabcolsep}{10pt}
\renewcommand{\arraystretch}{1}
        \centering
        \renewcommand{\arraystretch}{1}
        \vspace{-5pt}
        \begin{tabular}{lrrr}
            \toprule
            \textbf{Type} & \textbf{Precision} & \textbf{Recall} & \textbf{F1 score}\\  \hline
            Name &  $0.62$ & $0.35$& $0.45$\\
            Location & $0.50$ &  $0.67$ & $0.57$ \\
            Date Time  & $0.70$ & $0.60$ & $0.65$\\
            Phone Number &$0.58$ &$0.70$ & $0.63$  \\
            \hline
        \end{tabular}
        \caption{Private information detection method evaluation as precision, recall, and F1 score.}
         \label{tab:PI_eval}
     \vspace{-5pt}

\end{table}
\paragraph{Types of Private Information}

\begin{figure*}[t]
    \centering
    \begin{minipage}{0.45\textwidth}
        \centering
        \includegraphics[width=0.85\linewidth]{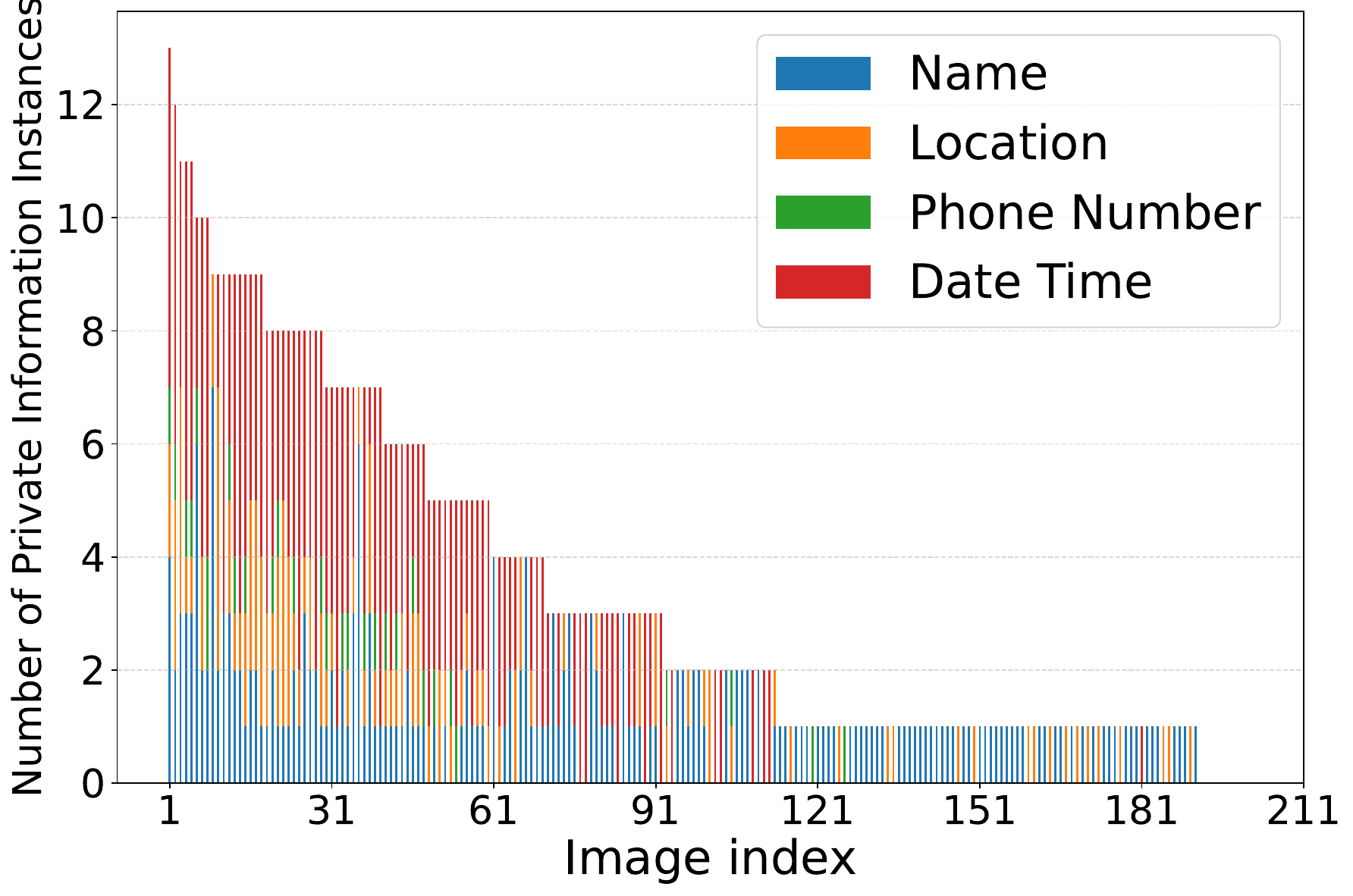}
        \caption{\label{fig:PI_dist} Number of private information detected in each pregnancy ultrasound within LAION-400M.}
    \end{minipage}
    \hfill 
    \begin{minipage}{0.45\textwidth}
        \centering
        \includegraphics[width=0.85\linewidth]{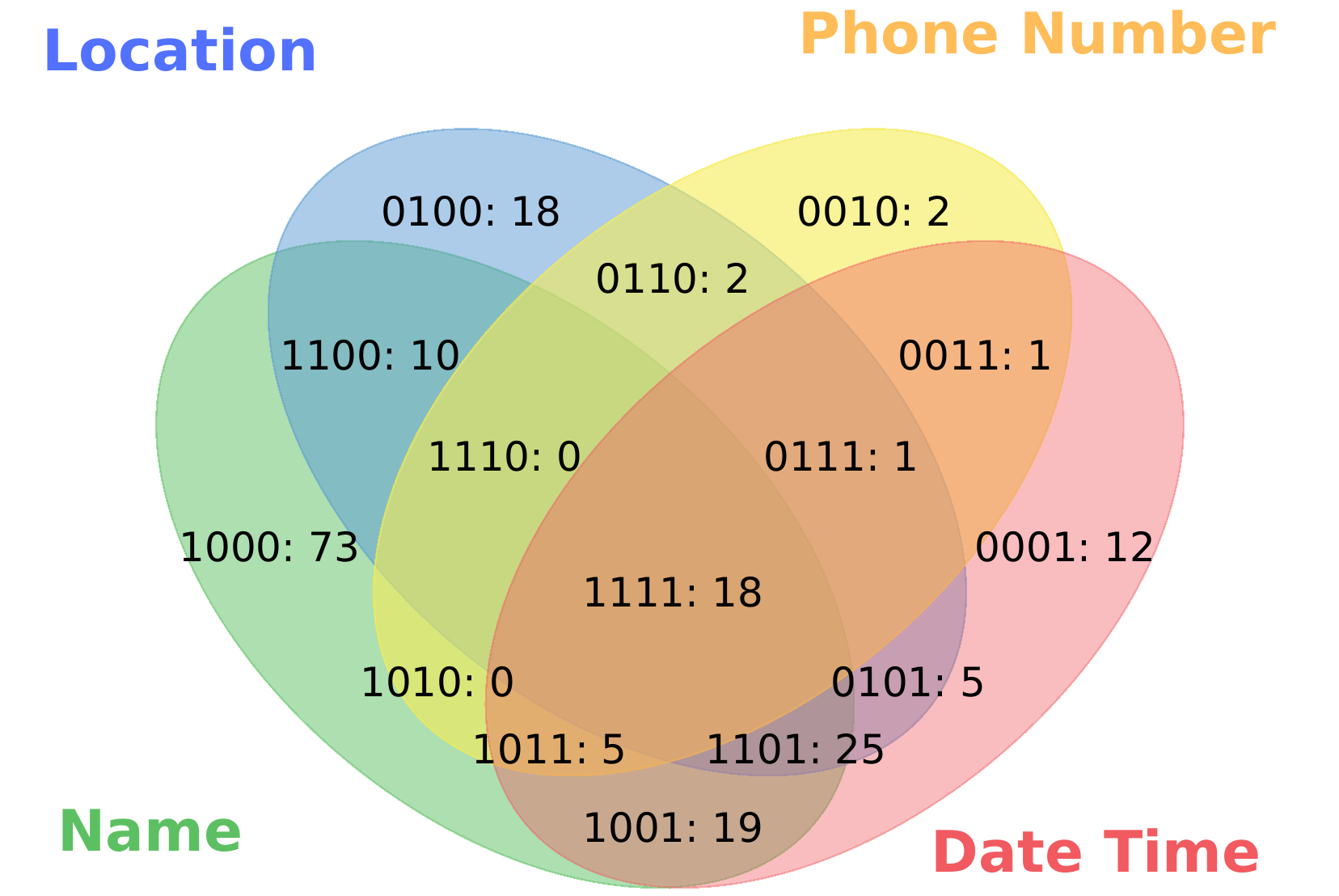}
        \caption{\label{fig:venn}Overlaps of private information types in the LAION-400M dataset. The four-digit binary code represents Name, Location, Phone Number, and Date Time, where ``1'' indicates presence and ``0'' indicates absence. }
    \vspace{-10pt}

    \end{minipage}
\end{figure*}
With the detected pregnancy ultrasound images in the LAION-400M dataset, we apply the methodology described previously to analyze the content and search for private information. Running Presidio on the 833 detected images, we find a total of 677 instances of private information. Table \ref{tab:PI_count} reports results both with and without duplicates (i.e., \textit{all images} and \textit{unique images}, respectively) to show how the raw, unmodified data behaves.
A large number of the detected instances are in the Date Time category, which is logical considering that ultrasound images typically include timestamps to document the exact time of the medical examination. Other prevalent private information categories detected in the pregnancy ultrasound images are Name and Location.

\paragraph{Private Information Identification Method Evaluation}

Relying solely on these counts provides limited insight, as they lack context regarding the accuracy and reliability of the method, as well as the potential errors that may arise across different components or modules.  To address this, we conduct a manual evaluation on a randomly selected subset of 200 pregnancy ultrasound images detected in the LAION-400M dataset. Two of the authors of the paper served as annotators to examine each of the 200 images, classifying the instances of private information within these images into four categories following the same definition as Presidio.
Each annotator reviewed the same set of 200 images, allowing for cross-checking and discussion to resolve uncertainties. Eventually, the annotations from each annotator are consolidated to create a unified ground truth.

We report precision, recall, and F1 score in Table \ref{tab:PI_eval} by comparing our private information identification method's detected strings to the ground truth from annotators, allowing for a margin of error to account for variations in string matching. Specifically, we consider two strings matched if their Levenshtein distance is less than 2 or the similarity is greater than 70\%. The Levenshtein distance measures how many single-character edits are needed to transform one word into another, which indicates the distance between two sequences. The similarity is calculated by taking twice the length of the longest common subsequence between two strings and dividing by the total number of characters in both strings. This flexibility was essential given the challenges of OCR on images with varying quality, which often resulted in incomplete or inaccurate text extraction. 

The results in Table \ref{tab:PI_eval} show that our method performs well in detecting Phone Number and Date Time information, as these primarily consist of numerical values and formatted strings, making them less susceptible to errors introduced by OCR. Conversely, recognizing Location and Name is more challenging. Manual inspection reveals that the framework exhibits difficulties in accurately extracting text associated with addresses. This is likely attributed to poor image resolution or distorted text rendering within the images. Similarly, off-the-shelf name recognition models displayed inconsistencies. For instance, the model could recognize \textit{Chole} but failed to identify \textit{Jessica}, despite both being common English names.
This implies that the model failed to recognize a significant number of actual Name instances. The low F1 scores suggest that the 228 detected Name instances from Table \ref{tab:PI_count} may not accurately represent the true number of Name instances in the dataset, with potentially more instances that have not been detected. To address these inaccuracies, instead of using a predefined recognizer, a more accurate and customized name recognizer may be necessary. 

\paragraph{Linked Instances of Private Information}
As discussed above, we detect various types of private information in the pregnancy ultrasound images in LAION-400M, often linking multiple details to specific individuals or events, elevating the potential risk of identity exposure. Some combinations of identifiers can be used to uniquely identify people, especially the mix of gender, birth date, and postal code, which can identify 87\% of individuals in the United States \citep{sweeney2000simple}. Our findings from Table \ref{tab:cluster_count} highlight that one of the largest clusters represents \textit{baby shower invitation cards}, which typically contain comprehensive details like names, locations, dates, and times. This is problematic because it aggregates and displays private data in a publicly accessible format, increasing the potential for misuse or misappropriation of private information and making individuals susceptible to privacy breaches and identity theft.

We analyze the linked instances of detected private information in Figures \ref{fig:PI_dist} and \ref{fig:venn}. Figure \ref{fig:PI_dist} shows the distribution of the number of private information instances detected per pregnancy ultrasound image. While the majority of the images have minimal or no private information, 86 images contain more than one type of sensitive information, with some images containing up to 13 distinct instances of private information. More specifically, Figure \ref{fig:venn} shows the intersections between the different types of private information per image. The results show that 22.9\% (191 images) contain at least one type of private information. Of these, 10.3\% (86 images) have more than one type of private information, and a smaller subset, 2.1\% (18 images), include all four types of private information. The most common overlap, occurring in 3\% (25 images), involves the combination of Name, Location, and Date Time information.

%% file: sec/5_conclusion.tex
\section{Recommendations}
Our work focuses on detecting private information in pregnancy ultrasound images within the LAION-400M dataset, representing only a fraction of the potential private data it may contain. There are other image types that could have private information, which we did not explore in this project. However, the focus on pregnancy ultrasound images is justified by their significant real-world privacy implications and the sensitivity of reproductive health data. Contrary to the view that this focus is narrow, we emphasize that half the population may undergo such ultrasounds. When including the fetus in these images, the privacy concern extends to virtually everyone at that stage of life, underscoring its broad relevance. Below, we outline several recommendations based on our findings.

\subsubsection*{Detection and De-Identification of Private Information} When creating a new dataset or utilizing a new dataset, it is crucial to make sure that individuals' privacy is preserved because the data will be used to train a model that might be vulnerable to membership attack, leaking the training data and sensitive information contained within the data \cite{carlini2021extracting, carlini2023extracting}. To address this similar issue, \citet{yu2021privacygan} applied neural network to detect private information and generative adversarial models to de-identify sensitive information such as facial features and car license plates on images collected from multimedia recording devices. This approach not only protects individuals' privacy, but also maintains the utility of the images. In similar manner, dataset creators or users should employ various methods, such as Presidio \cite{mendels2018presidio} or neural networks, to detect and de-identify private information. Developing systematic automated tool that can perform these tasks is essential to establish privacy-preserving dataset.

\subsubsection*{Practice Consent}Consent is the best tool for limiting excessive collection of personal data and respecting individual autonomy \cite{Froomkin_2019}. Informed consent is necessary to protect individuals and uphold their rights. Without it, people often worry their information might be sold or analyzed by AI for marketing purposes, leading to feelings of rights violations
\citep{Andreotta_Kirkham_Rizzi_2021}. The internet offers a vast pool of data, but collecting data from it may violate people's rights. We should strive to collect datasets that respect individual rights by ensuring people are well-informed and offering them the option to opt-out, rather than assuming consent by default, especially in machine learning where data is used in diverse applications. Obtaining consent from both data providers and owners fosters ethical and transparent practices within the field and more effectively aligns their goals and expectations. 

\subsubsection*{Privacy-Preserving Training} Given our findings on private information in image datasets that may be memorized by models \cite{ju2025watch}, we recommend robust privacy-preserving methods during training in addition to preprocessing datasets. Differential Privacy (DP) protects individual data by adding noise to query results or representations while preserving statistical utility \cite{Dwork2006DifferentialP}. It has been applied across different domains, such as facial recognition \citep{CHAMIKARA2020privacyface}, image generation \citep{yu2021privacygan}, and medical images classification \cite{wu2019p3sgd}. However, our discovery of 531 duplicate images (38.9\% of retrieved data) challenges DP's core assumption that adjacent datasets differ by only one record. Duplicates allow an individual’s data to persist even after removal, undermining DP guarantees and highlighting the critical role of careful data collection in privacy-preserving systems.

Beyond DP, other methods include a distributed selective SGD framework, where multiple models collaboratively learn from their datasets without directly sharing them \cite{Shokri2015privacydl}. Moreover, the concept of knowledge transfer through a teacher-student model, where the teacher model is trained on disjoint data and the student model learns from the teacher's noisy aggregated responses, illustrated innovative strategies to protect privacy during the training phase \cite{papernot2016semi, papernot2018scalable, liu2020revisit}. By integrating these privacy-preserving techniques, we can improve the confidentiality of the data while maintaining the integrity and utility of the training process.

\subsubsection*{Further Research}Applying Nissenbaum's theory of contextual integrity, we bring attention to the issues that arise when data that is shared in one context and with a certain intention is then scraped and used without consent in a domain and for purposes other than those originally intended. At the same time, the contextual nature of what should be considered private makes it difficult to create clear guarantees that privacy will be respected. In the case of pregnancy ultrasounds, their dual medical and social functions make them especially sensitive data, and their out-of-context use and the privacy issues which result from it are of particular concern. There is an urgent need for attention to issues at the intersection of reproductive health, privacy, and data management. We aim to draw needed attention to these issues and to stimulate social and academic debate to clarify the relevant norms and build consensus around the appropriate handling of such data.

\section{Conclusion}

We explored the presence of private information within the LAION-400M dataset, focusing specifically on the retrieval of pregnancy ultrasound images. We employed both retrieval-based and classifier-based approaches to identify relevant images within the dataset. The images we found are not only typical clinical ultrasound images but also included baby shower invitations, pregnancy announcements, images of ultrasound machines, and ultrasound photos within frames. These findings underscore the diverse contexts in which ultrasound images are shared, reflecting real-world uses. We found presence of private information in those images, with many instances where multiple types of personal data co-occur within a single image, which increases the risk of identity exposure. 

We believe that this work represents a significant step toward privacy in image datasets, shedding light on the prevalence of embedded private information and highlighting the importance of open-source data auditing for a safe and responsible machine learning community. 

%% file: sec/6_ethic_statement.tex
\section{Ethical Statement}
We recognize the ethical implications inherent in our research. We are committed to upholding the privacy of individuals, particularly in cases where data may be collected inadvertently from the internet without explicit consent. Our project involves the creation of a dataset that includes pregnancy ultrasound images, which contain highly sensitive personal information. To protect these data, we will take the following measures: the dataset will not be shared publicly, and all images and corresponding trained models will be securely deleted upon the completion of the project. We take these steps to ensure the highest respect for individual privacy and data integrity.

%% file: sec/7_adverse_impact.tex
\section{Adverse Impact Statement}
While the main goal of our work is to identify and highlight the presence of private information in publicly available image datasets, we are aware of the dual-use risk. Specifically, malicious actors could adapt the proposed methodology to extract private information from other datasets and use it to harm or threaten individuals through identity theft, harassment, targeted misinformation, or other violations of privacy \cite{king2024rethink}. These risks point to a serious issue: many large datasets contain personal and identifiable information, often collected through data scraping without proper protection or consent \cite{birhane2021multimodal}. 

We hope our work draws attention to these concerns and encourages the research community to approach this space with caution and to develop ethical frameworks, auditing mechanisms, and clear guidelines for dataset collection, release, and downstream use. Future work might also explore technical safeguards to help mitigate potential harms from existing datasets.

%% file: sec/8_acknowledgement.tex
\section{Acknowledgements}
This work was supported by JSPS KAKENHI No. JP23H00497 and JP22K12091, JST CREST Grant No.~JPMJCR20D3, JST FOREST Grant No.~JPMJFR216O, and The University of Osaka IDS Co-Creation Project Na22990007.